\documentclass[11pt,a4paper]{article}

\usepackage[utf8]{inputenc}
\usepackage[T1]{fontenc}
\usepackage{lmodern}
\usepackage{geometry}
\usepackage{float}
\usepackage{placeins}
\usepackage{subcaption}

\usepackage{amsmath, amssymb, amsfonts}
\usepackage{bm}

\usepackage{graphicx}
\usepackage{caption}
\usepackage{subcaption}
\usepackage{booktabs}
\usepackage{multirow}
\usepackage{array}
\usepackage{longtable}
\usepackage{longtable}

\usepackage[numbers,sort&compress]{natbib}

\usepackage[hidelinks]{hyperref}

\usepackage{setspace}
\title{\textbf{Leakage-Aware Bandgap Prediction on the JARVIS-DFT Dataset:\\
A Phase-Wise Feature Analysis}}

\author{
Gaurav Kumar Sharma\\
\small Metallic Materials Technology\\
\small Faculty of Materials Science and Technology\\
\small TU Bergakademie Freiberg, Germany\\
\small \texttt{gaurav-kumar.sharma@student.tu-freiberg.de}
}

\date{}
\begin{document}
\maketitle 
\begin{abstract}
In this study, we perform a systematic analysis of the JARVIS-DFT bandgap dataset
and identify and remove descriptors that may inadvertently encode band-structure
information, such as effective masses. This process yields a curated,
leakage-controlled subset of 2,280 materials. Using this dataset, a three-phase
modeling framework was implemented that incrementally incorporated basic physical
descriptors, engineered features, and compositional attributes. The results show
that tree-based models achieve $R^2$ values of approximately 0.88--0.90 across all
phases, indicating that expanding the descriptor space does not substantially
improve predictive accuracy when leakage is controlled. SHAP analysis
consistently identified the dielectric tensor components as the dominant
contributors. This work provides a curated dataset and baseline performance
metrics for future leakage-aware bandgap prediction studies.
\end{abstract}
\section{Introduction}

The electronic bandgap is a fundamental material property that governs charge
transport, optical absorption, and carrier excitation, thereby determining the
suitability of materials for applications in optoelectronics, photovoltaics, and
power electronics \cite{ref1,ref2}. Accurate bandgap prediction is therefore a
central problem in materials science and device engineering. Although density
functional theory (DFT) has been the dominant computational tool for estimating
bandgaps, standard exchange--correlation approximations are known to
systematically underestimate experimental values, and more accurate approaches,
such as hybrid functionals or many-body perturbation methods, remain
computationally expensive for large-scale screening.

The growing availability of high-throughput DFT repositories has enabled the use
of machine learning (ML) models as efficient surrogate predictors for bandgaps.
Databases such as the Joint Automated Repository for Various Integrated
Simulations (JARVIS) provide large, curated datasets of electronic and structural
properties, making them attractive benchmarks for materials informatics research
\cite{ref3}. Leveraging such repositories, numerous studies have demonstrated
that ML models can reproduce DFT-level bandgaps with reasonable accuracy while
dramatically reducing the computational cost. Beyond large databases, several
groups have also shown that classical ML models, when combined with careful
feature selection and physically motivated descriptors, can perform well even on
smaller or curated datasets \cite{ref4,ref5,ref6,ref7}.

Despite this progress, an important methodological issue has often received
insufficient attention: feature leakage. Feature leakage occurs when input
descriptors unintentionally encode information that is too closely related to the
target property, leading to an artificially inflated model performance and
misleading conclusions about the predictive capability \cite{ref8}. In materials
datasets, this risk arises naturally because many commonly used descriptors are
derived from DFT electronic structure calculations and are therefore physically
correlated with the underlying band structure \cite{ref3,ref9}. Such correlations
do not imply data leakage by default; however, if descriptors that strongly depend
on electronic eigenvalues or related quantities are included without careful
screening, the resulting models may exploit hidden shortcuts rather than learning
transferable structure--property relationships. Models trained under these
conditions cannot be reliably used for realistic predictions or materials
discovery tasks \cite{ref8,ref10}.

In this work, we adopt a systematic strategy to address this challenge. We begin
by constructing a leakage-free subset of the JARVIS bandgap database through
domain-aware screening of candidate descriptors, ensuring that no input feature
unfairly encodes the target bandgap. Using this curated dataset as a foundation,
we introduce a three-phase modeling framework designed to isolate the impact of
feature complexity on predictive performance. Phase~I employs only simple,
physically interpretable descriptors to establish a transparent baseline. In
Phase~II, engineered quantities are introduced to evaluate whether they provide
genuinely independent information. Phase~III incorporates composition-based
descriptors generated using \texttt{Matminer} to examine how model performance
evolves with a richer but potentially riskier feature space \cite{ref9}.

To further validate model reliability, we employ SHAP-based interpretability
analysis across all three phases. This enables examination of how individual
features contribute to predictions and verification of whether model behavior is
governed by physically meaningful trends or by hidden correlations that may
signal residual leakage \cite{ref12}. Rather than treating interpretability as a
post-hoc visualization tool, it is used here as an integral component of model
validation.

Overall, this study aims to provide (i) a clean, leakage-free baseline dataset for
bandgap prediction derived from JARVIS, (ii) a controlled, phase-wise assessment
of how increasing feature complexity affects model accuracy, and (iii)
physically interpretable insights into the most influential descriptors through
SHAP analysis. By explicitly addressing feature leakage within a
material-specific context, this study seeks to improve the reliability and
interpretability of ML-based bandgap prediction models.

\section{Method}

\subsection{Data Preprocessing and Curation}

The dataset used in this study was constructed by combining two JARVIS
repositories, \texttt{JARVIS-dft\_3d} and \texttt{JARVIS-dft\_3d\_2021}. Both
datasets originated from the same optB88-vdW workflow, ensuring descriptor
compatibility. After merging them through their shared descriptor fields, the
combined dataset contained 131{,}716 structures with 31 features, with
\texttt{optb88vdw\_bandgap} selected as the target property \cite{ref3}. Duplicate
compositions were resolved by retaining only the lowest-energy polymorphs,
yielding 51{,}905 unique entries. Placeholder strings such as \texttt{na},
\texttt{[]}, \texttt{None}, and \texttt{nan} were standardized to \texttt{NaN} for
consistent pre-processing. To address sparsity in several JARVIS descriptors,
filtering was performed using a core set of physical quantities essential for
meaningful bandgap modeling: formation energy per atom, density, number of atoms,
dimensionality, space-group number, bulk and shear moduli, Poisson ratio,
dielectric tensor components (\texttt{epsx}, \texttt{epsy}, \texttt{epsz}),
convex-hull energy, and the bandgap. Rows missing any of these descriptors were
removed, resulting in 34{,}969 entries.

To ensure that the dataset remained physically meaningful, a few additional
filters were applied. First, entries with positive formation energies were
removed, as these phases are thermodynamically unstable \cite{ref12}. For
Poisson's ratio, a working range of 0.1--0.5 was used. Although classical
isotropic elasticity places the lower bound of Poisson's ratio near 0.2
\cite{ref13}, experimental measurements and first-principles studies have shown
that several anisotropic and low-dimensional materials can exhibit stable
Poisson's ratios approaching 0.1 without violating mechanical stability
\cite{ref14,ref15}. This slightly relaxed lower limit helped avoid rejecting such
materials unnecessarily. Extremely large elastic constants were also removed, as
bulk moduli above $\sim$300~GPa or shear moduli above $\sim$200~GPa frequently
arise from numerical instabilities, convergence issues, or misreported
structures in large high-throughput databases, rather than representing reliable
mechanical behavior for the majority of entries \cite{ref16}. Similarly, entries
with anomalous dielectric tensor components were excluded to ensure a physically
meaningful linear-response behavior in the dataset. Finally, materials with
invalid space-group indices were discarded. After applying these checks, the
dataset was reduced to 9{,}459 structures. To avoid mixing materials with
fundamentally different dimensionalities and bonding behaviors, 0D clusters, 1D
systems, and intercalated structures were removed, leaving 5{,}355 entries
categorized as either 2D or 3D. A binary descriptor (\texttt{is\_3D}) was defined
accordingly. The \texttt{max\_efg} descriptor, often missing in JARVIS, was
reconstructed when the full EFG tensor was available by taking the maximum
absolute tensor component across all the atomic sites. Entries in which
reconstruction was not possible were discarded.

Leakage testing was performed using the electron and hole effective masses, since
these quantities encode band-structure curvature and can therefore act as
potential leakage carriers. For this diagnostic step, only materials with
complete \texttt{avg\_elec\_mass} and \texttt{avg\_hole\_mass} values were
retained, yielding a temporary subset of approximately 1{,}915 entries. Leakage
signals were detected for both descriptors; therefore, they were removed entirely
from the feature space. Since the modeling pipeline does not use these
descriptors, the full curated set of 2{,}280 materials was kept for the
subsequent feature-engineering and machine-learning phases. The final dataset
consisted of 2{,}280 physically valid materials with 23 complete descriptors and
no missing values, forming the basis of a three-phase modeling framework.

\subsection{Statistical Analysis for Checking Data Integrity}

Several electronic and optical features, such as effective masses, SLME,
spillage, and electronic dielectric components, were removed because their
limited availability would have eliminated nearly 50--60\% of the 2{,}280
entries. Features such as \texttt{encut} and \texttt{kpoint\_length\_unit} were
also excluded since they reflect DFT numerical settings rather than intrinsic
material properties and are known sources of leakage.

To maintain physical relevance, only descriptors that were complete and
scientifically meaningful were retained. After reducing the dataset from
51{,}905 unique entries to 2{,}280, a statistical integrity check was carried out
to ensure that the curated dataset remained reliable and representative.

The assessment focused on whether (i) key property distributions remained
meaningful, (ii) the chemical space was preserved, and (iii) the core descriptor
ranges remained physically reasonable. Overall, these checks confirmed that
strict filtering did not introduce unwanted bias or remove the essential
variability required for machine learning models.

\subsubsection{Distribution Consistency and Chemical-Space Coverage}

A Kolmogorov--Smirnov (K--S) test comparing the bandgap distributions before and
after filtering showed a significant shift ($D = 0.0562$, $p < 0.001$), which is
expected after removing incomplete or unstable entries. Even so, the overall
shape of the distribution remained similar, and the curated dataset is now
dominated by physically well-behaved compounds. PCA on 12 complete descriptors
further showed that the top five components captured 77.4\% of the variance of
the raw subset, and nine components were required to reach 95\%. This indicates
that, despite the large reduction in size, the essential chemical and
structural diversity was reasonably preserved.

\subsubsection{Physical-Property Range Preservation}

To evaluate whether essential physical variability was maintained, the ranges of
three fundamental descriptors were compared between the raw and curated datasets
in Table~\ref{tab:range_preservation}.

\begin{table}[H]
\centering
\caption{Fraction of physical-property ranges preserved after dataset curation,
comparing the raw and leakage-controlled JARVIS-DFT subsets.}
\label{tab:range_preservation}
\begin{tabular}{lc}
\toprule
Property & Range Preserved (\%) \\
\midrule
Bandgap & 38.5 \\
Density & 91.5 \\
Formation energy per atom & 44.8 \\
\bottomrule
\end{tabular}
\end{table}
Density shows excellent retention, while the narrowing of the bandgap and
formation energy ranges reflects the removal of unstable or incomplete entries,
an expected trade-off when restricting the dataset to physically meaningful
values.
\subsubsection{Summary and Interpretation}

The combined results from the K--S test, PCA, and property range comparison show
that the final dataset remains statistically robust and chemically diverse,
even after major size reduction. Although some rare chemistries and extreme
values were lost, which is unavoidable when enforcing physical consistency and
removing leakage-prone descriptors, the curated dataset retains the variability
needed for developing reliable, leakage-free machine-learning models and is
scientifically suitable for further predictive work.

\subsection{Data Leakage Detection and Mitigation Framework}

Machine learning models are particularly vulnerable to data leakage, which
occurs when the features in the input dataset inadvertently encode the target
property. This can result in phantom progress, where benchmark metrics improve
without corresponding gains in real-world generalization, thereby undermining
scientific reliability in material machine learning \cite{ref10}. At the feature
level, such misleading performance gains often arise when proxy variables encode
the target outcome, producing artificially inflated accuracy without genuine
predictive insights \cite{ref8}.

In materials datasets, this risk is amplified by the use of computationally
derived descriptors and metadata, which can introduce strong but unintended
correlations if not carefully screened. To address these challenges, this study
adopts a systematic leakage detection and mitigation framework. In addition,
redundancy-aware feature-selection strategies, such as the relevance redundancy
optimization employed in the MODNet framework, are essential for improving
robustness and generalization in data-limited materials science applications
\cite{ref6}.

A three-stage protocol was implemented to detect and mitigate the target leakage.

\subsubsection{Feature Risk Categorization}

The features were categorized based on their physical origin and potential to
encode bandgap information to create a shortlist for formal leakage testing
(Table~\ref{tab:feature_risk}).
\begin{table}[H]
\centering
\caption{Categorization of descriptors based on leakage risk, physical origin,
and potential to encode bandgap-related information.}
\label{tab:feature_risk}
\begin{tabular}{p{3cm} p{6cm} p{6cm}}
\toprule
Risk Category & Descriptors & Rationale \& Origin \\
\midrule
High-Risk &
\texttt{avg\_elec\_mass}, \texttt{avg\_hole\_mass} &
Derived from band-structure curvature (DFT post-processing); high potential to
encode target. \\

Medium-Risk &
\texttt{epsx}, \texttt{epsy}, \texttt{epsz} &
Dielectric tensor from DFPT (independent perturbation); physical correlation
without mathematical encoding of bandgap. \\

Low-Risk &
\texttt{formation\_energy\_per\_atom}, \texttt{density}, \texttt{nat},
\texttt{bulk\_modulus\_kv}, \texttt{shear\_modulus\_gv}, \texttt{poisson},
\texttt{ehull}, \texttt{max\_efg}, \texttt{spg\_number}, \texttt{is\_3D},
\texttt{dimensionality} &
Intrinsic structural and thermodynamic quantities; no electronic band-structure
information. \\
\bottomrule
\end{tabular}
\end{table}
\subsubsection{Baseline Performance Estimation}

A leakage-free performance baseline was established using only low- and
medium-risk descriptors. Random Forest and XGBoost models were trained on this
feature set, deliberately excluding the high-risk \texttt{avg\_elec\_mass} and
\texttt{avg\_hole\_mass} descriptors.

\subsubsection{Incremental Feature-Impact Analysis}

Each high-risk feature was individually added to the baseline set, and the
models were retrained. The evaluation included three configurations: baseline,
baseline + \texttt{avg\_elec\_mass}, and baseline + \texttt{avg\_hole\_mass}. A
feature was flagged as a source of leakage if its inclusion caused an
unexpected and physically unrealistic performance improvement. The addition of
either effective mass caused $R^2$ to surge to 0.90--0.93 and MAE to fall to
approximately 0.05~eV, markedly exceeding the typical DFT bandgap prediction
benchmarks and confirming both as leakage features.
\subsection{Machine Learning}

\subsubsection{Hierarchical Feature Engineering Framework}

To study how different types of descriptors affect the bandgap predictions, the
feature set was constructed in three stages. The first stage included only
fundamental physical descriptors. In the second stage, a group of engineered
features derived from the basic physical descriptors was added to the model. In
the final stage, composition-based descriptors that capture broader chemical
trends were introduced. Constructing the feature space in this hierarchical
manner enabled a clearer assessment of how the model behavior evolved as
additional information was incorporated, rather than mixing all descriptor
types from the outset.

\paragraph{Phase I: Fundamental Physical Descriptors}

Phase I established a baseline model using 12 intrinsic physical properties
directly extracted from the curated JARVIS-DFT dataset. These descriptors capture
core thermodynamic, mechanical, electronic, and structural characteristics known
to influence bandgap behavior.

\begin{itemize}
\item \textbf{Thermodynamic:} \texttt{formation\_energy\_per\_atom},
\texttt{ehull}
\item \textbf{Mechanical:} \texttt{bulk\_modulus\_kv},
\texttt{shear\_modulus\_gv}, \texttt{poisson}
\item \textbf{Electronic:} dielectric tensor components
(\texttt{epsx}, \texttt{epsy}, \texttt{epsz})
\item \textbf{Structural:} \texttt{density}, \texttt{nat}, \texttt{is\_3D}
\item \textbf{Local electronic environment:} \texttt{max\_efg}
\end{itemize}

This phase forms the benchmark for the later stages. Notably, in these
experiments, this set already captured most of the model’s predictive ability.
Consequently, the feature additions in Phases~II and~III did not lead to
noticeable improvements over this baseline. This makes Phase~I a useful point
of comparison for evaluating whether more complex descriptors provide any real
benefit.

\paragraph{Phase II: Engineered Physical Descriptors}

Phase II expands the feature set by adding a set of engineered descriptors that
capture relationships that are not directly available from the raw Phase~I
features. These descriptors were derived through simple analytical
transformations of the basic physical properties, with a small stabilizing
constant ($\varepsilon = 1 \times 10^{-12}$) applied when needed to avoid
numerical issues. The engineered features used in this phase were as follows:

\begin{itemize}
\item \textbf{Dielectric response:} \texttt{dielectric\_mean},
\texttt{dielectric\_anisotropy}
\item \textbf{Ductility indicator:} \texttt{pugh\_ratio} (\texttt{gv/kv})
\item \textbf{Elastic wave and stiffness proxies:} \texttt{v\_t\_proxy},
\texttt{v\_l\_proxy}, \texttt{specific\_stiffness}
\item \textbf{Coupling descriptor:} \texttt{stability\_stiffness\_ratio}
\end{itemize}

The goal of this phase was to determine whether introducing these physically
motivated combinations improved the model before adding the chemistry-driven
descriptors used in Phase~III.

\paragraph{Phase III: Compositional and Orbital Descriptors}

Phase III introduces compositional and orbital features by adding them directly
to the fundamental descriptors from Phase~I. This isolates the impact of
chemical information from the engineered physical transformations in Phase~II.
Here, the goal is to combine physical, chemical, and orbital features into the
input feature set to evaluate their impact on prediction performance.
Extraction of features such as element properties, valence orbital information,
and ionic properties was performed using the \texttt{Matminer} toolkit
\cite{ref9} through the composition of the materials in the data.

A total of 132 MagpieData features were added to the base data. In addition to
the Magpie atomic features, orbital and ionic property features were also
extracted. In addition, 11 engineered chemical and structural descriptors were
generated, as shown in Table~\ref{tab:engineered_descriptors}.
\begin{table}[H]
\centering
\caption{Engineered chemical, orbital, and structural descriptors introduced in
Phase~III, including their mathematical definitions, physical interpretations,
and value ranges within the dataset.}
\label{tab:engineered_descriptors}
\begin{tabular}{p{4cm} p{4cm} p{6cm} p{3cm}}
\toprule
Descriptor Name & Mathematical Definition & Physical Meaning & Range (Dataset) \\
\midrule
Bond Polarity Index &
$(\Delta \chi)^2$ &
Measures electronegativity contrast; higher values indicate increased ionic
character &
0.00--8.29 \\

Atomic Size Homogeneity &
$1 / \left(1 + (r_{\max} - r_{\min}) / r_{\mathrm{mean}}\right)$ &
Uniformity of atomic radii; 1 = perfectly uniform &
0.32--1.00 \\

Relative Electronegativity Range &
$(\chi_{\max} - \chi_{\min}) / \chi_{\mathrm{mean}}$ &
Normalized spread in electronegativity among constituent atoms &
0.00--1.34 \\

Radius Mismatch &
$(r_{\max} - r_{\min}) / r_{\mathrm{mean}}$ &
Degree of mismatch between smallest and largest atomic sizes &
0.00--2.15 \\

Atomic Size Uniformity &
$r_{\mathrm{mean}} / r_{\max}$ &
Ratio capturing size compactness relative to the largest atom &
0.38--1.00 \\

Radius Variance &
$(r_{\mathrm{std}})^2$ &
Statistical variance of covalent radii &
0.00--7056.00 \\

Pauling Ionicity Proxy &
$1 - \exp\!\left[-0.25 \cdot (\Delta \chi)^2\right]$ &
Pauling's ionicity measure capturing ionic contribution to bonding &
0.00--0.87 \\

d-Hybridization Tendency &
$f_d$ &
Extent of d-orbital participation in bonding &
0.00--0.97 \\

p--d Orbital Interaction Index &
$f_p \times f_d$ &
Strength of hybridization between p- and d-orbitals &
0.00--0.175 \\

s--p Promotion Index (clipped) &
$f_s / f_p$ &
Relative contribution of s vs.\ p valence electrons &
0.38--10.00 \\

Transition-Metal Electron Index &
$\mathrm{mean}(f_d)$ &
Characterizes transition-metal content in the composition &
0.00--0.97 \\
\bottomrule
\end{tabular}
\end{table}
After the addition of the engineered and Magpie features, the total number of
features increased to 170. Variance and correlation filtering were applied to
identify redundant inputs. A variance threshold of 0.001 was used, which removed
one near-constant feature. This was followed by correlation filtering with a
cutoff value of 0.95. A conservative threshold was intentionally adopted to avoid
removing chemically meaningful descriptors; this step eliminated 46 highly
correlated features from the dataset. In total, 47 features were removed,
resulting in a cleaned dataset with 123 features. Also, during variance and
correlation filtering, one dielectric tensor component was removed due to strong
collinearity with the remaining components, reflecting statistical redundancy
rather than physical exclusion.

After this preprocessing, feature selection was performed using the
XGBoost-based importance ranking. The models were then trained using feature
subsets ranging from 10 to 123 features, with the subset size increasing in steps
of five to determine the optimal number of features for achieving the highest
prediction accuracy.
\subsection{Machine Learning Modeling}

Three training permutations were performed for each machine-learning model to
test the robustness of the model predictions. Five different algorithms were
used: Ridge Regression, SVR, Random Forest, XGBoost, and CatBoost. These models
were chosen because they cover a spectrum from linear to kernel-based to
tree-based learning methods. This helps in understanding how the material
datasets behave across different types of algorithms.

\subsubsection{Model Architectures and Hyperparameter Optimization}

Ridge Regression was used as the linear baseline. It was implemented through a
simple \texttt{StandardScaler--Ridge} pipeline, and the regularization strength
was kept fixed at $\alpha = 1.0$ for all experiments. For the nonlinear baseline,
Support Vector Regression (SVR) was used with an RBF kernel.

The main SVR hyperparameters were tuned using a 5-fold \texttt{GridSearchCV}.
The following search space was employed:
\begin{itemize}
\item $C$: 1, 10, 100, 500
\item $\varepsilon$: 0.01, 0.05, 0.1
\item $\gamma$: \texttt{scale}, 0.1, 0.5, 1
\end{itemize}

These two models were also tested alongside three tree-based methods: Random
Forest, XGBoost, and CatBoost. These algorithms are well known for handling
nonlinear patterns and feature interactions that frequently occur in materials
datasets. Table~\ref{tab:tree_hyperparameters} provides the precise values of the
hyperparameters and settings of the tree models.

Three tree-based models were included because they can capture nonlinear
relationships and feature interactions more effectively than linear or kernel
methods \cite{ref17}. The specific hyperparameters and model settings used for
these tree models are presented in Table~\ref{tab:tree_hyperparameters}.

\begin{longtable}{p{3cm} p{3cm} p{8cm}}
\caption{Hyperparameter configurations used for tree-based machine-learning
models (Random Forest, XGBoost, and CatBoost) under conservative, balanced, and
aggressive settings.}
\label{tab:tree_hyperparameters} \\
\toprule
Model & Configuration & Hyperparameters \\
\midrule
\endfirsthead

\toprule
Model & Configuration & Hyperparameters \\
\midrule
\endhead

\midrule
\multicolumn{3}{r}{\textit{Continued on next page}}
\endfoot

\bottomrule
\endlastfoot

Random Forest & Conservative &
\texttt{n\_estimators = 500}, \texttt{max\_depth = 13},
\texttt{min\_samples\_leaf = 5} \\

& Balanced &
\texttt{n\_estimators = 600}, \texttt{max\_depth = None} \\

& Aggressive &
\texttt{n\_estimators = 700}, \texttt{max\_depth = None},
\texttt{min\_samples\_leaf = 1} \\

XGBoost & Conservative &
\texttt{n\_estimators = 500}, \texttt{learning\_rate = 0.05},
\texttt{max\_depth = 6} \\

& Balanced &
\texttt{n\_estimators = 600}, \texttt{learning\_rate = 0.10},
\texttt{max\_depth = 8} \\

& Aggressive &
\texttt{n\_estimators = 700}, \texttt{learning\_rate = 0.30},
\texttt{max\_depth = 6} \\

CatBoost & Conservative &
\texttt{iterations = 1000}, \texttt{learning\_rate = 0.01},
\texttt{depth = 6}, \texttt{l2\_leaf\_reg = 5} \\

& Balanced &
\texttt{iterations = 3000}, \texttt{learning\_rate = 0.05},
\texttt{depth = 10}, \texttt{l2\_leaf\_reg = 1} \\

& Aggressive &
\texttt{iterations = 2000}, \texttt{learning\_rate = 0.03},
\texttt{depth = 8}, \texttt{l2\_leaf\_reg = 3} \\

\end{longtable}
\subsubsection{Training and Evaluation Protocol}

The models were trained and tested across the three phases using a single,
consistently defined train--test split. A 4:1 ratio was used for training and
testing, which is common practice in materials informatics. Keeping the split
fixed across all phases made it possible to directly compare how the
predictions changed when new feature sets were introduced. The hyperparameters
were also kept constant within each model type to allow a fair phase-wise
comparison. For Ridge and Support Vector Regression (SVR), feature scaling was
performed using a standard scaler.

To prevent data leakage, the scaler was fitted only to the training data and
then applied to the test data. The tree-based models were trained on raw,
unscaled features. A random seed of 42 was used for all the randomization steps
to maintain reproducibility. The evaluation settings and metrics remained the
same for all phases and models to ensure valid comparison. Model evaluation was
performed in two parts:

\begin{itemize}
\item \textbf{Numerical Performance Evaluation:} The following metrics were used
to quantify model performance.
\begin{itemize}
\item \textbf{$R^2$ score:} Measures how closely the predicted bandgap values
match the actual bandgaps.
\item \textbf{Mean Absolute Error (MAE):} Represents the average absolute
difference between predicted and true bandgaps (in eV).
\item \textbf{Mean Squared Error (MSE):} Similar to MAE but applies a squared
penalty to larger errors, making the metric more sensitive to large deviations.
\end{itemize}
Together, these three metrics provide a comprehensive view of each model's
accuracy and error characteristics.

\item \textbf{SHAP-Based Model Interpretation:} Shapley additive explanation
(SHAP) analysis \cite{ref11} was used to interpret the model outputs and
understand how individual features contribute to the bandgap prediction. This
analysis was applied only to the tree-based models in each phase, as the SHAP
values for tree ensembles are well-established and more reliable. The SHAP
study offers phase-wise insights into which features have the greatest
influence on the predictions.
\end{itemize}
\section{Results}

\subsection{Effect of Effective-Mass Descriptors on Model Performance}

\begin{figure}[htbp]
\centering

% ---------- First row ----------
\begin{subfigure}[b]{0.49\linewidth}
\centering
\includegraphics[width=\linewidth]{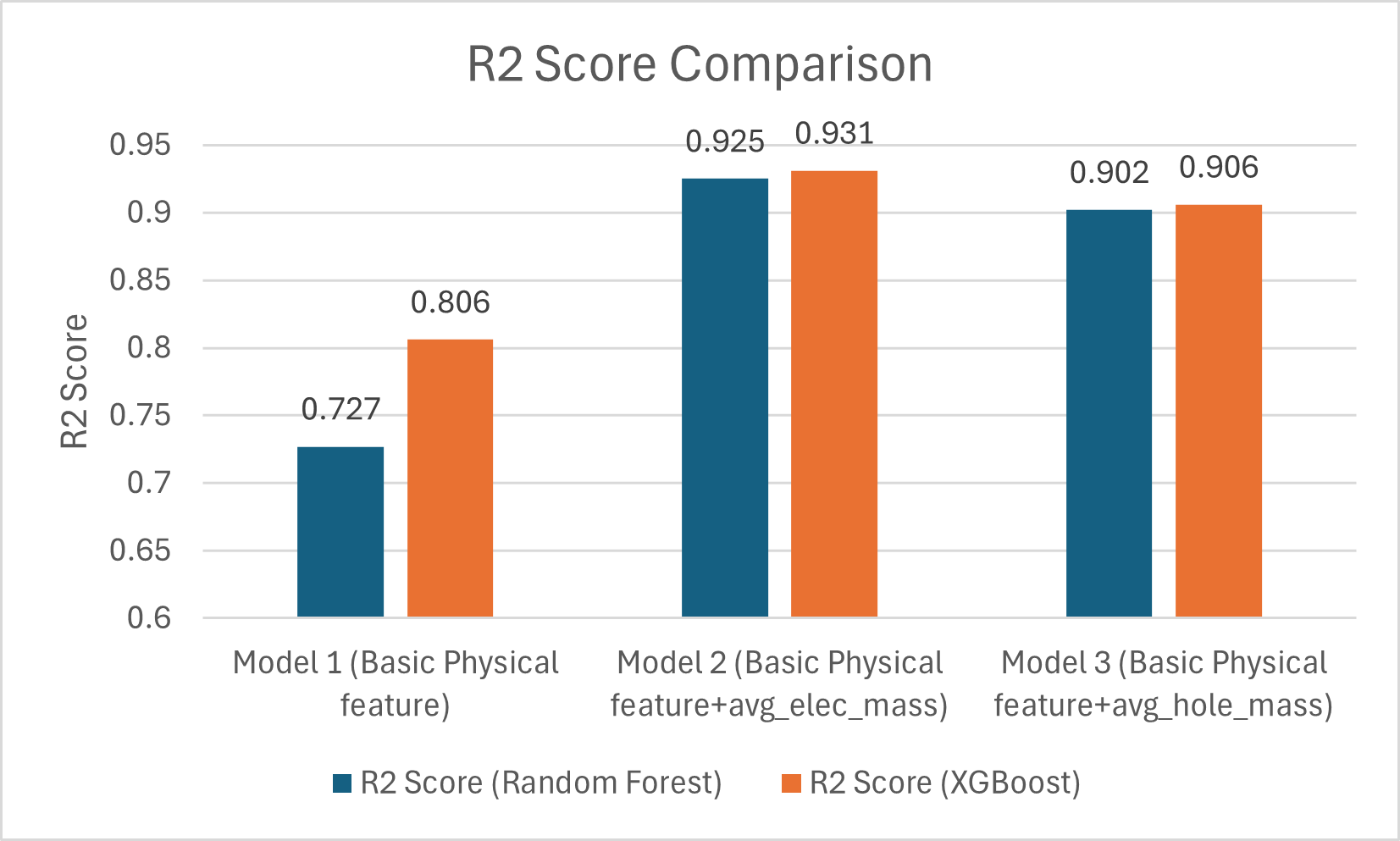}
\caption{$R^2$ score}
\label{fig:r2}
\end{subfigure}
\hfill
\begin{subfigure}[b]{0.49\linewidth}
\centering
\includegraphics[width=\linewidth]{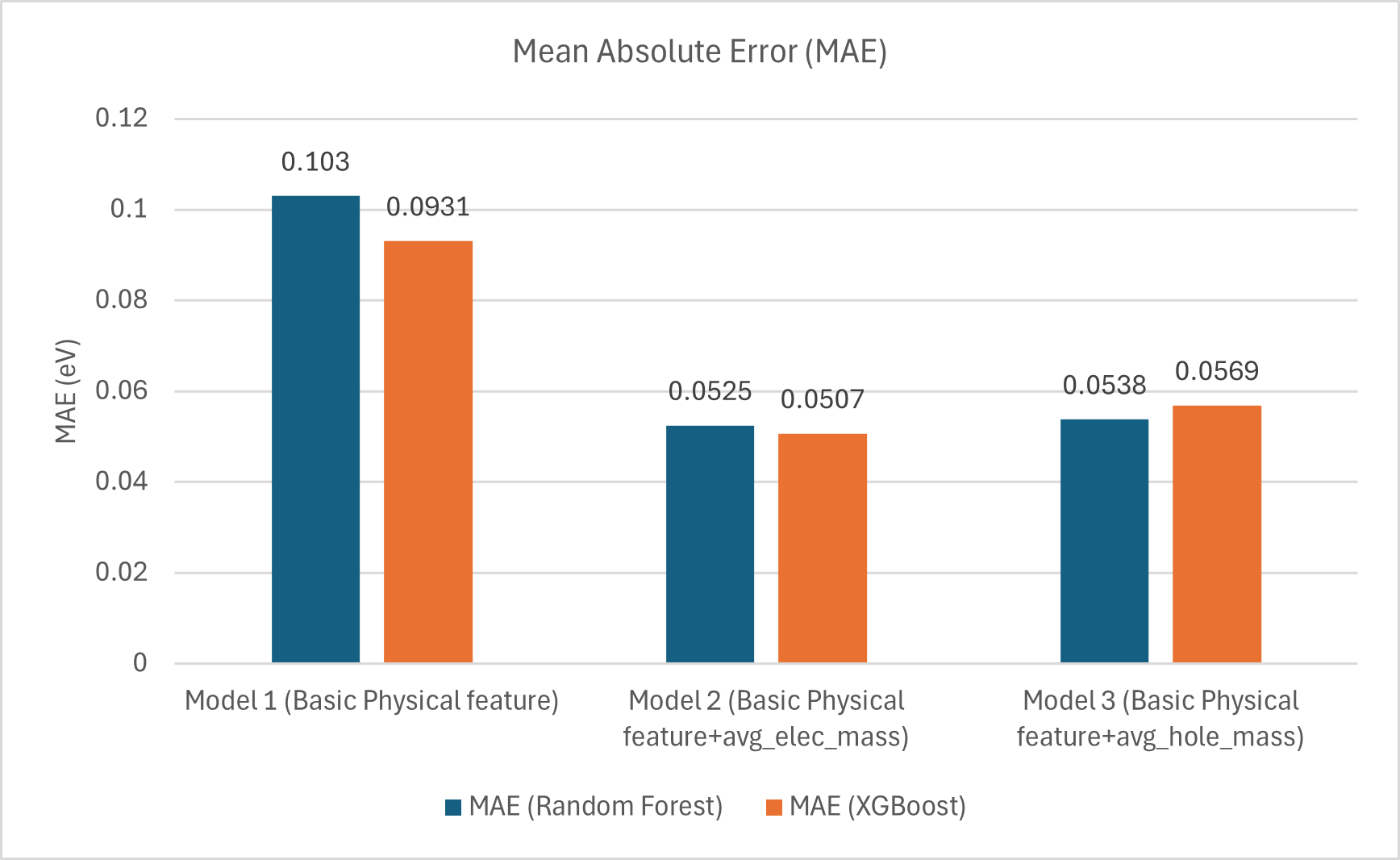}
\caption{MAE}
\label{fig:mae}
\end{subfigure}

\vspace{0.6em}

% ---------- Second row ----------
\begin{subfigure}[b]{0.62\linewidth}
\centering
\includegraphics[width=\linewidth]{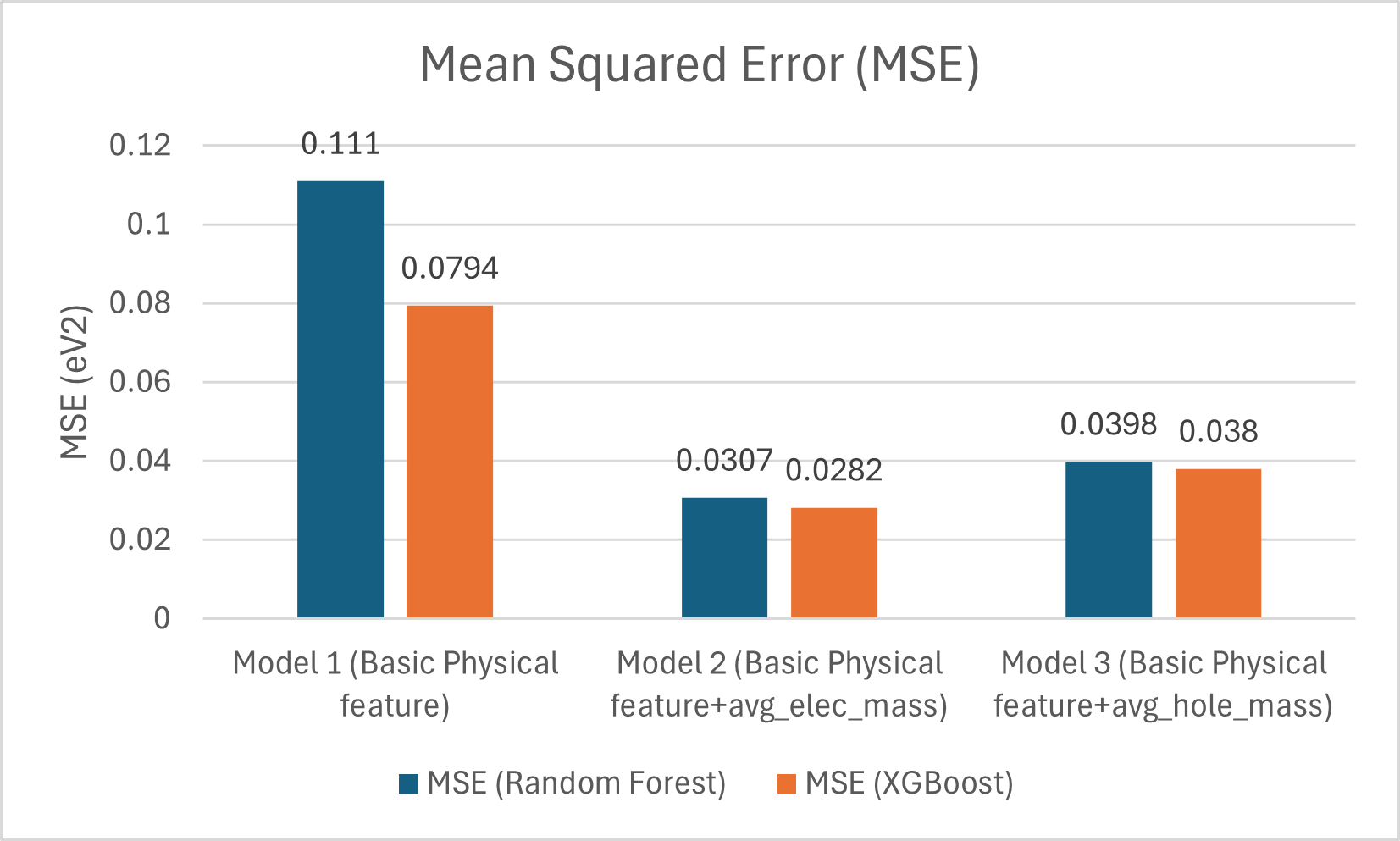}
\caption{MSE}
\label{fig:mse}
\end{subfigure}

\caption{Effect of effective-mass descriptors on model performance.
(a) $R^2$ score, (b) mean absolute error (MAE), and (c) mean squared error (MSE)
for models trained with and without effective-mass features.}
\label{fig:effective_mass_comparison}
\end{figure}

Including the descriptors \texttt{avg\_elec\_mass} and \texttt{avg\_hole\_mass}
led to a substantial increase in the model accuracy. The $R^2$ score increased
from 0.727--0.806 to 0.902--0.931, while the MAE decreased from approximately
0.10~eV to 0.055~eV and the MSE decreased from 0.08--0.10 to 0.028--0.039. The
corresponding performance comparison is presented in
Figure~\ref{fig:effective_mass_comparison}.

\subsection{Phase-wise Model Performance Across Feature Sets}

The three-phase framework for assessing model prediction accuracy with variations
in the input features is outlined here. In each phase, five models were trained
to evaluate the performance of the feature set. Phase-wise modeling also allows
a clear understanding of how feature engineering affects the prediction when the
same input features are used and how the incorporation of Matminer’s atomic features
together with engineered chemical and structural descriptors influences bandgap
prediction in materials. 
\begin{table}[htbp]
\centering
\caption{Phase-wise model performance for bandgap prediction. Reported values
correspond to the best configuration among the three hyperparameter settings
for each model.}
\label{tab:phasewise_results}
\begin{tabular}{llccc}
\toprule
Model & Phase & $R^2$ & MAE (eV) & MSE (eV$^2$) \\
\midrule
Ridge & I: Fundamental & 0.458 & 0.497 & 0.615 \\
      & II: Engineered & 0.460 & 0.490 & 0.613 \\
      & III: Compositional (Full features) & 0.599 & 0.466 & 0.455 \\
\midrule
SVR   & I: Fundamental & 0.751 & 0.236 & 0.283 \\
      & II: Engineered & 0.713 & 0.261 & 0.325 \\
      & III: Compositional (Full features) & 0.755 & 0.252 & 0.278 \\
\midrule
Random Forest
      & I: Fundamental & 0.899 & 0.135 & 0.115 \\
      & II: Engineered & 0.904 & 0.129 & 0.106 \\
      & III: Compositional (Top 110 features) & 0.880 & 0.134 & 0.137 \\
\midrule
XGBoost
      & I: Fundamental & 0.894 & 0.131 & 0.119 \\
      & II: Engineered & 0.889 & 0.131 & 0.126 \\
      & III: Compositional (Top 110 features) & 0.893 & 0.133 & 0.121 \\
\midrule
CatBoost
      & I: Fundamental & 0.887 & 0.135 & 0.127 \\
      & II: Engineered & 0.883 & 0.150 & 0.133 \\
      & III: Compositional (Top 110 features) & 0.896 & 0.137 & 0.117 \\
\bottomrule
\end{tabular}
\end{table}
\subsection{Phase-wise Analysis of Model Performance}

The three-phase framework for assessing model prediction accuracy with variations
in the input features is outlined here. In each phase, five models were trained to
evaluate the performance of the feature set. Phase-wise modeling also allows a clear
understanding of how feature engineering affects the prediction when the same input
features are used and how the incorporation of Matminer’s atomic features together
with engineered chemical and structural descriptors influences bandgap prediction
in materials. Table~\ref{tab:phasewise_results} presents the model-wise results
(best of the three hyperparameter settings) for the predicted bandgap values for the
three phases.

\subsubsection*{Phase I: Baseline Physical Descriptors Provide Foundational Predictive Performance}
Phase I involved predictions using a selected baseline feature set comprising only
physically measurable descriptors. The cleaned input data were sourced from the
JARVIS dataset, and five machine learning models were trained on this uniform feature
set. The observed results align well with the known behaviors of these algorithms.
As expected, Ridge Regression exhibited the weakest performance, failing to achieve
an $R^2$ above 0.50, indicating that linear models underfit the underlying relationships.
Support Vector Regression (SVR) displayed moderate predictive capability, whereas
tree-based models performed strongly, with XGBoost achieving the highest accuracy,
as shown in Table~\ref{tab:phasewise_results}.

\subsubsection*{Phase II: Engineered Physical Descriptors Added to the Baseline Feature Set}
In Phase II, the baseline physical descriptors were supplemented with a set of
engineered features derived from them, creating a slightly larger but still
physics-motivated input space. The models were retrained using the same train--test
split as in Phase I to ensure a fair comparison.

Overall, the expanded feature set did not lead to any noticeable improvements.
Ridge Regression shows only a marginal change (from $R^2 = 0.458$ to $0.460$), while
SVR actually performs slightly worse. The tree-based models, Random Forest,
XGBoost, and CatBoost, exhibited essentially unchanged performance between Phase I
and Phase II, with differences in $R^2$ limited to $0.003$--$0.01$. These changes are
numerically negligible relative to the overall model accuracy and indicate that the
addition of engineered physical descriptors does not provide any meaningful
performance gain for the ensemble-based methods. Taken together, the Phase II
results indicate that adding engineered physical descriptors to the baseline feature
set does not materially enhance the predictive performance of any model. The overall
trend remains essentially identical to that observed in Phase I.

\subsubsection*{Phase III: Matminer Features with Baseline Descriptors}
In Phase III, the feature set was expanded quite a bit by adding a larger group of
Matminer descriptors on top of the original physical features. These include several
compositional, chemical, and orbital features, along with a handful of
Matminer-engineered features already listed in the Methods section. No engineered
physical descriptors from Phase II were used here.

\begin{figure}[htbp]
\centering

\begin{minipage}[t]{0.48\linewidth}
\centering
\includegraphics[width=\linewidth]{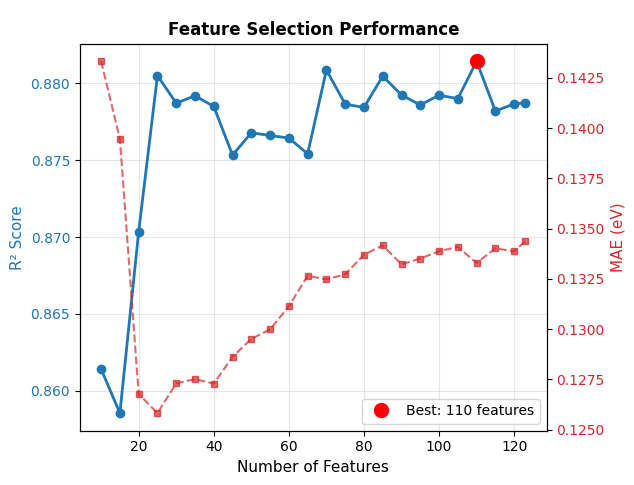}
\captionof{figure}{Model performance as a function of the number of input features
used to identify the optimal feature subset size.}
\label{fig:feature_selection}
\end{minipage}
\hfill
\begin{minipage}[t]{0.48\linewidth}
\centering
\includegraphics[width=\linewidth]{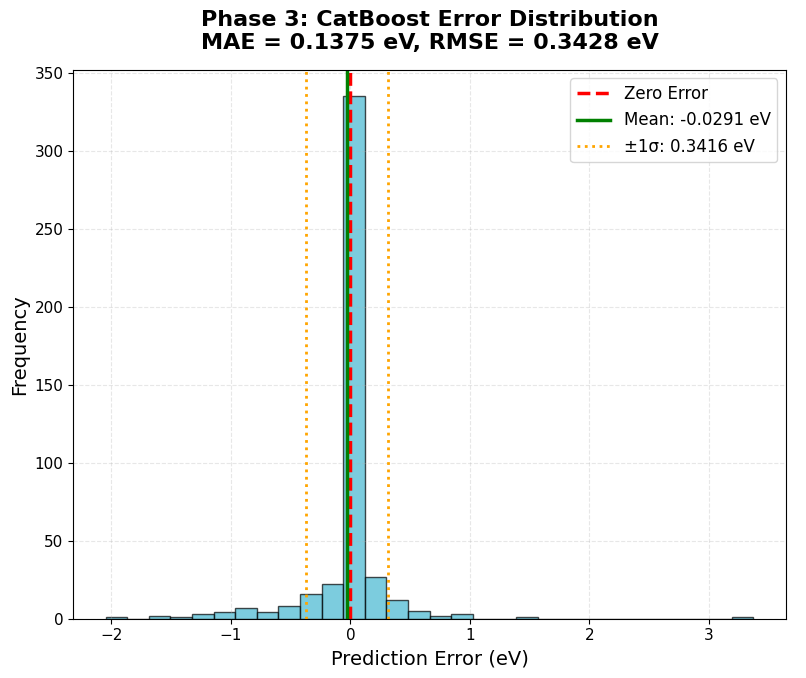}
\captionof{figure}{Error distribution for the CatBoost model trained using the
Phase III feature set, showing the distribution of prediction residuals.}
\label{fig:error_distribution}
\end{minipage}

\end{figure}

Before training the main models, a small feature sweep was performed using a simple
Random Forest model. The goal was just to check how the accuracy changes as the
number of features increases. Feature counts from 10 up to 123 were tested in steps
of five. The trend peaked around 110 features, as shown in
Figure~\ref{fig:feature_selection}, so that number was taken forward for the
tree-based models in this phase. With those 110 features, the ensemble models were
retrained. Random Forest reached an $R^2$ of 0.880 with an MAE of 0.134~eV and an MSE
of 0.137~eV$^2$. XGBoost performed slightly better, recording $R^2 = 0.893$, MAE =
0.133~eV, and MSE = 0.121~eV$^2$. Among all the models tested in this phase, CatBoost
yielded the highest score, with $R^2 = 0.896$, MAE = 0.137~eV, and MSE =
0.117~eV$^2$.

Ridge Regression and SVR were trained using the full set of 123 features. The ridge
improved compared to the previous phases and reached $R^2 = 0.599$, with an MAE of
0.466~eV and an MSE of 0.455~eV$^2$. SVR stayed close to its earlier performance,
giving $R^2 = 0.755$, MAE = 0.252~eV, and MSE = 0.278~eV$^2$.

An error-distribution plot (Figure~\ref{fig:error_distribution}) was generated for
CatBoost, as it was the best-performing model. The mean prediction error was
0.029~eV, and the standard deviation was approximately 0.3416~eV. Most predictions
were close to zero, and only a few noticeable outliers appeared in the distribution.

\subsection{Phase-Wise Parity Plot Comparison}

\begin{figure}[H]
\centering

\begin{subfigure}[b]{0.55\linewidth}
\centering
\includegraphics[width=\linewidth]{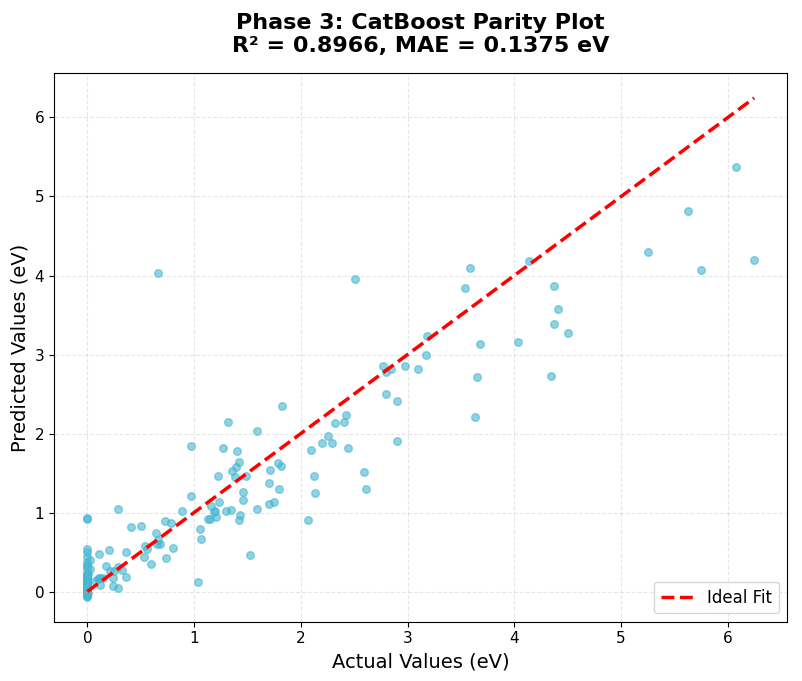}
\caption{Phase III}
\end{subfigure}

\caption{Phase-wise parity plots for the best-performing models in Phase I,
Phase II, and Phase III, comparing predicted and reference bandgap values.
The predicted values align closely with the ideal-fit line across all phases,
indicating consistent predictive behavior despite increasing feature complexity.}
\label{fig:parity_plots}
\end{figure}

Figure~\ref{fig:parity_plots} shows the parity plots of the best-performing models
from each phase. The three plots exhibit a very similar overall pattern: the
predicted values align closely with the ideal-fit line, and the spread of points
remains nearly unchanged across the phases. Despite the progressive increase in
feature dimensionality, the general prediction behavior remained consistent, with
no substantial shift in accuracy or scatter characteristics from Phase I to
Phase III.
\begin{figure}[H]
\centering

\begin{subfigure}[b]{0.48\linewidth}
\centering
\includegraphics[width=\linewidth]{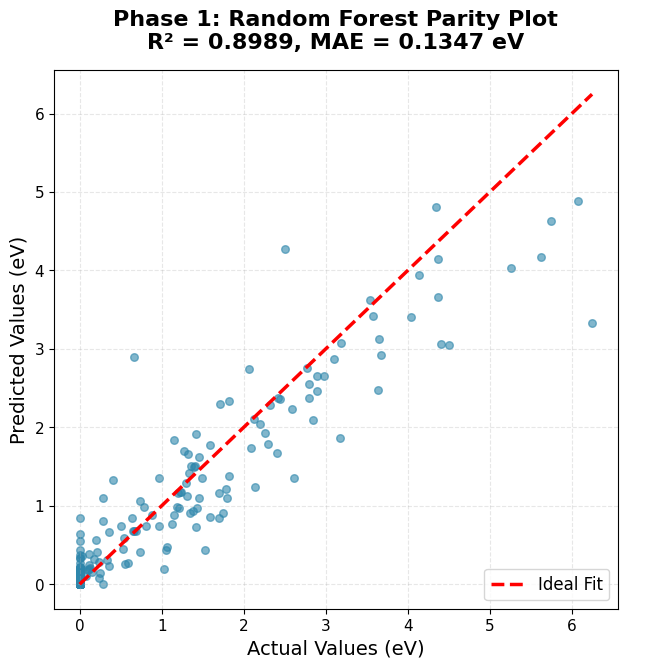}
\caption{Phase I}
\end{subfigure}
\hfill
\begin{subfigure}[b]{0.48\linewidth}
\centering
\includegraphics[width=\linewidth]{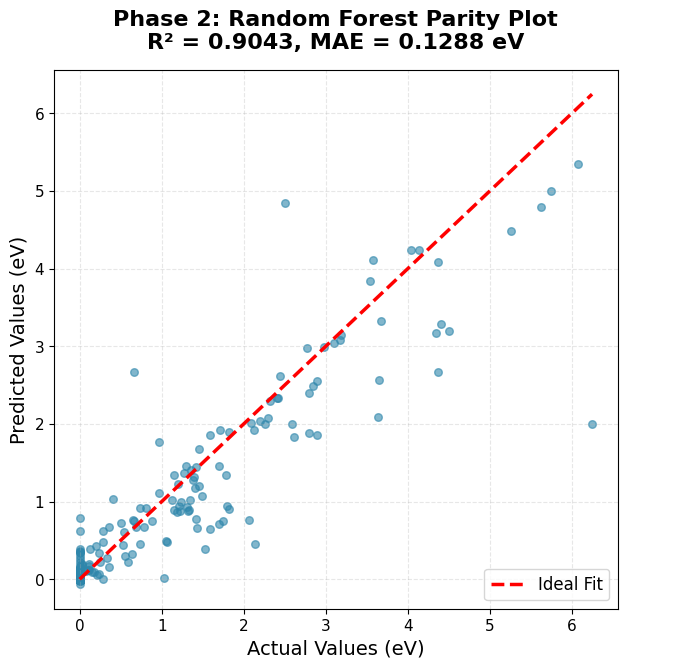}
\caption{Phase II}
\end{subfigure}

\end{figure}
\subsection{Feature Importance Across the Three-Phase Modeling Framework (SHAP Analysis)}

\begin{figure}[H]
\centering

\includegraphics[
  width=0.72\linewidth,
  height=0.22\textheight,
  keepaspectratio
]{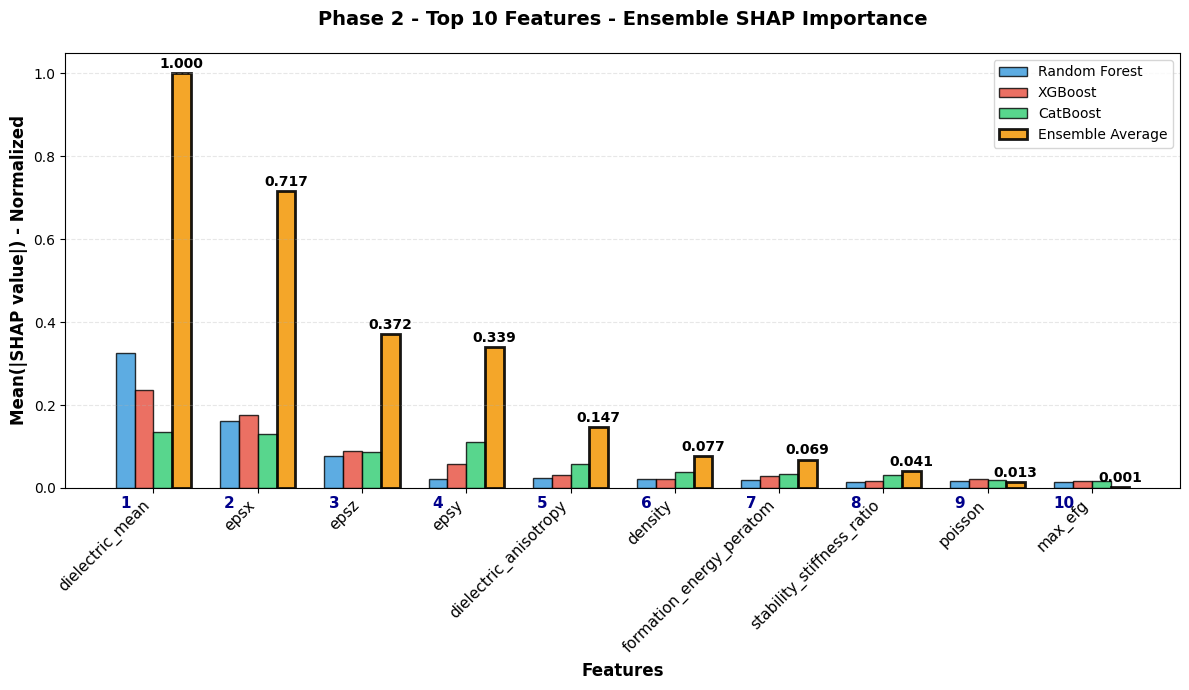}
\caption*{(b) Phase II: Baseline physical descriptors + Engineered physical descriptors}

\vspace{0.6em}

\includegraphics[
  width=0.72\linewidth,
  height=0.22\textheight,
  keepaspectratio
]{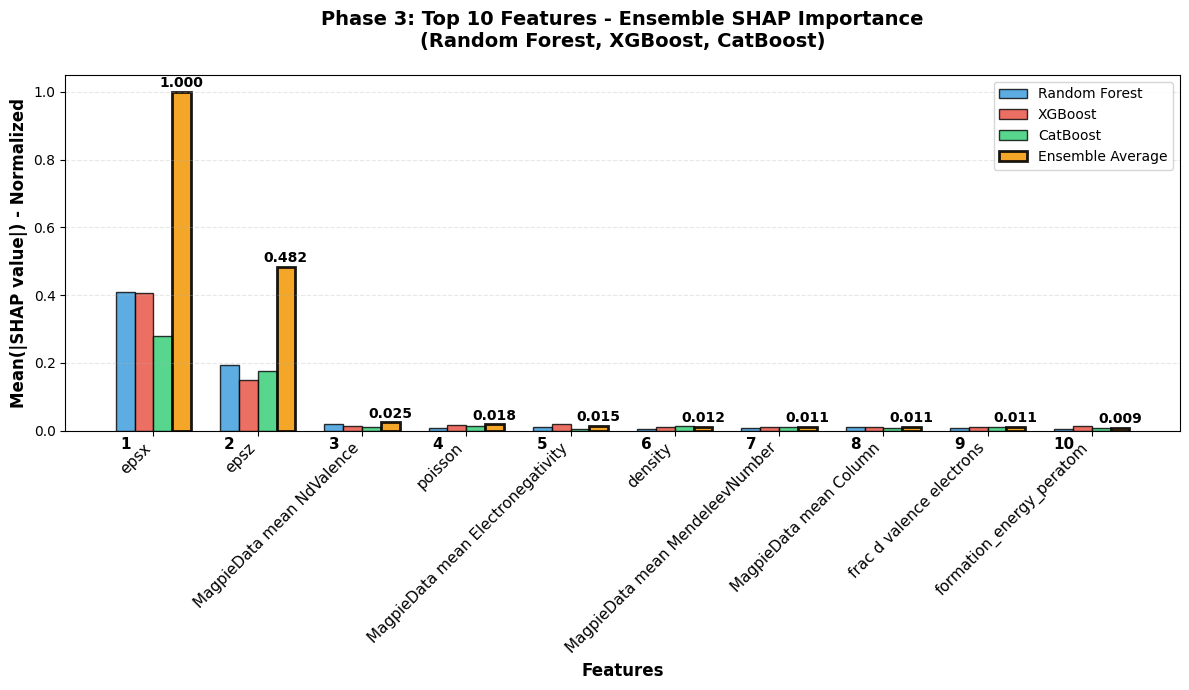}
\caption*{(c) Phase III: Fundamental physical descriptors + Compositional and Matminer descriptors}

\caption{SHAP-based feature importance analysis for the best-performing models
across the three modeling phases.
(a) Phase I: Baseline physical descriptors.
(b) Phase II: Baseline physical descriptors with engineered features.
(c) Phase III: Baseline physical descriptors with compositional and Matminer features.
Dielectric tensor components consistently dominate the feature rankings, while the added compositional and Matminer-derived descriptors contribute primarily as secondary features.}
\label{fig:shap_importance}
\end{figure}

Across all three phases, the SHAP analysis (Figure~\ref{fig:shap_importance}) shows a
consistent pattern: the dielectric tensor components remain the dominant predictors
of the bandgap, while the remaining features mainly provide secondary support to the
model. In Phase I, the three principal dielectric components (epsx, epsy, and epsz)
clearly dominated the feature rankings. In Phase II, the engineered dielectric
descriptor introduced in this phase emerged as the strongest contributor, with the
individual dielectric tensor components following closely behind. In Phase III,
where Matminer-derived compositional and chemical attributes are added, the dielectric
features again appear at the top of the importance list, whereas the Magpie
descriptors contribute primarily as supporting features rather than leading ones.
This phase-wise consistency highlights that the dielectric response of a material
plays a central role in driving the predictive performance of the models, regardless
of how the overall feature space evolves.

\enlargethispage{2\baselineskip}

\begin{figure}[H]
\centering
\includegraphics[
  width=0.80\linewidth,
  height=0.22\textheight,
  keepaspectratio
]{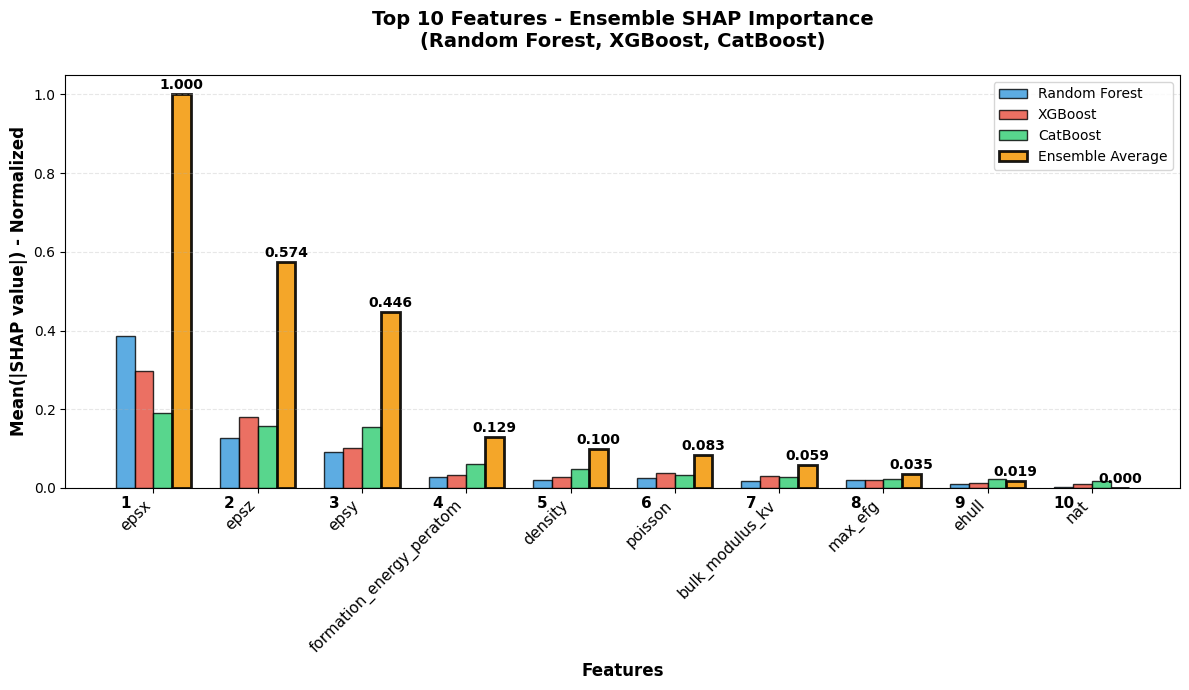}
\end{figure}

\noindent\textbf{(a) Phase I: Baseline physical descriptors.}

\section{Discussion}

\subsection{Importance of Leakage-Free Modeling and Comparison with Prior Work}

The performance comparison in Figure~\ref{fig:effective_mass_comparison} highlights
the strong influence of effective-mass descriptors on the predictive accuracy of
bandgap regression models. When the models were trained using only the physical
descriptors, the accuracy remained within a stable and physically reasonable range.
However, introducing the features \textit{avg\_elec\_mass} and
\textit{avg\_hole\_mass} produced an immediate and disproportionate improvement
across all evaluation metrics, with $R^2$ increasing to above 0.90 and MAE and MSE
decreasing by approximately a factor of two. Such abrupt changes suggest that these
descriptors introduce target leakage into the modeling pipeline.

The underlying reason for this behavior can be traced to the physics of band-structure
calculations. In semiconductor theory, the electron and hole effective masses are
defined through the curvature of the energy bands near the extrema,
\begin{equation}
m^{*} = \hbar^{2} \left( \frac{d^{2}E}{dk^{2}} \right)^{-1}
\end{equation}
This relationship is well established in the literature~\cite{ref1,ref18}. The JARVIS-DFT workflow reports these quantities as part of the band-structure outputs~\cite{ref3}. Because the bandgap and effective masses originate from the same underlying electronic structure calculations, the effective mass descriptors inherently encode information related to the target property. Their presence therefore enables the models to exploit correlations that arise directly from the band-structure itself, rather than inferring the bandgap purely from independent physical descriptors.
While previous studies have highlighted the risk of unintended target proxies and feature-related pitfalls in materials machine learning workflows~\cite{ref7}, few studies have explicitly quantified how band-structure-derived descriptors can distort the predictive performance in bandgap regression. The present analysis directly contrasts the leakage-free results with those obtained using effective-mass terms, making the magnitude of their impact transparent. This comparison underscores the necessity of excluding such descriptors when constructing generalizable bandgap-prediction models. By removing the effective-mass features from all phases of the modeling, this study established a physically grounded and leakage-controlled baseline for evaluating the machine learning performance on the JARVIS-DFT dataset.
\subsection{Hierarchical Feature Engineering and Evidence of Saturation in Predictive Accuracy}

The three-phase feature expansion framework was designed to examine whether
progressively richer descriptors yielded measurable improvements in bandgap
prediction. However, the results show that the $R^2$ values remain confined to a
narrow band of approximately 0.88--0.90 across all phases, suggesting that the
predictive capacity of leakage-free features in the JARVIS-DFT dataset saturates
early. The models did not exhibit the sequential accuracy gains typically expected
when meaningful new information was added to the feature space.

In Phase I, the classical models trained on fundamental physical descriptors already
achieved strong predictive performance. These descriptors include the dielectric
tensor components, elastic moduli, Poisson’s ratio, formation energy per atom, and
material dimensionality. SHAP analysis consistently identifies the dielectric tensor
components as the dominant contributors, reflecting their strong influence on
dielectric screening and band-edge behavior, two physical factors closely tied to a
material’s bandgap.

Phase II introduces engineered descriptors designed to capture more nuanced physical
phenomena, such as dielectric anisotropy, mean dielectric response, Pugh ratio, and
stability--stiffness relationships. Although these features enrich the physical
interpretability of the model, their contribution to the accuracy remains modest.
The small variations in $R^2$, MAE, and MSE were numerically minor relative to the
overall model accuracy, indicating that these engineered descriptors provided
incremental rather than transformative information relative to the Phase I feature
set.

Phase III incorporates a much broader set of Matminer-derived chemical and orbital
descriptors, substantially expanding the dimensionality of descriptor space. Despite
this increase, the overall performance remained comparable to the earlier phases,
again centered around $R^2 \approx 0.89$. The SHAP rankings show that the dielectric
tensor components continue to dominate, whereas the Matminer descriptors primarily
act as supporting features. This apparent saturation likely reflects the limited information content of
optB88vdW-level descriptors with respect to bandgap variability, rather than a
fundamental limitation of machine-learning models themselves. These findings suggest that simply increasing the number
of descriptors does not necessarily enhance the predictive performance; even with
the addition of an extensive suite of Matminer features, the model does not exhibit
appreciable gains in accuracy.

Taken together, these observations point to feature-information saturation within
the JARVIS-DFT dataset for classical machine-learning models. Expanding the descriptor
richness from fundamental physical descriptors to engineered physical features and
finally to comprehensive Matminer compositional descriptors did not produce successive
improvements in predictive capability. Instead, the models converged to a stable
accuracy ceiling governed primarily by the dielectric properties. It should also be
noted that this conclusion applies specifically to the classical ML models examined
here; deep learning architectures were not evaluated and may display different
sensitivities to high-dimensional feature spaces or latent representation learning.

\subsection{SHAP Interpretation Across Phases: Consistent Dominance of Dielectric Features}

SHAP analysis provides a unifying perspective on how the models predict across all
three phases. In Phase I, the dielectric tensor components emerged as the most
influential descriptors for every tree-based model, far outweighing the mechanical
or structural features. This behavior is physically intuitive because the dielectric
response is closely connected to how a material screens electric fields and reflects
the electronic polarizability near the band edges, both of which correlate strongly
with bandgap trends across the chemical space.

Phase II maintained this pattern. Although the introduction of engineered descriptors
improves interpretability, it does not alter the hierarchy of the feature importance.
The engineered dielectric descriptors become more prominent in this phase, but they
reinforce the same underlying signal captured by the raw dielectric components rather
than replacing it.

Even in Phase III, where the descriptor space expands substantially through the
inclusion of Matminer chemical and orbital features, the dielectric tensor components
remain the dominant contributors. The compositional and chemical features play
meaningful but clearly secondary roles, supporting rather than redefining the
predictive structure of the models. The SHAP rankings show that the tree-based models
consistently rely on dielectric behavior as the primary source of information, with
mechanical and chemical descriptors providing additional refinements.

Taken together, the SHAP analysis highlights a robust and phase-independent pattern:
the dielectric properties govern the predictive performance across all feature sets,
whereas the chemical, structural, and mechanical descriptors function mainly as
complementary inputs. This reinforces that the models are learning physically grounded
relationships rather than shortcut correlations. It also supports the broader
conclusion that the JARVIS bandgap dataset contains a strong dominant
signal---dielectric screening---that classical models capture effectively even when
additional descriptors are introduced.

\subsection{Limitations and Future Work}

This study has several limitations that also point toward clear directions for future
development.

First, the dataset size decreased substantially during the curation stage. Although
strict filtering ensured physical consistency, it also removed many chemically complex
and compositionally diverse materials. The resulting dataset was more homogeneous and
may have limited the model’s exposure to rare chemistries, thereby constraining the
generalizability of the predictions.

Second, the models were trained exclusively on optB88vdW bandgaps computed within the
JARVIS-DFT workflow. As a result, the predicted values inherit the systematic behavior
and biases of this specific functional and do not directly translate to higher-fidelity
hybrid-functional (e.g., HSE06) or experimental bandgaps. The lack of high-fidelity
electronic-structure descriptors, many of which are lost during the leakage removal
process, further limits the richness of the predictive space.

Third, this study relies solely on classical machine-learning models. While these
methods perform well on moderately sized datasets, they may be insufficient to capture
broader, highly nonlinear electronic-structure correlations that could emerge from
larger, more heterogeneous datasets or from deep-learning architectures specifically
designed to learn long-range chemical and structural patterns.

These limitations suggest several concrete pathways for future research.
(1) Aggregating data from multiple repositories, such as the Materials Project, OQMD,
and AFLOW, would expand the dataset, restore chemical diversity, and incorporate
bandgaps computed using a range of exchange--correlation functionals.
(2) Once larger datasets are available, deep learning models can be developed to learn
more expressive latent representations that classical models may not capture.
(3) Transfer learning offers a promising strategy: a deep model can first be trained on
large, low-fidelity datasets (e.g., optB88vdW) and subsequently fine-tuned on smaller
sets of high-fidelity HSE or carefully curated experimental bandgaps. This approach
may help bridge the gap between DFT-level predictions and experimental reality.

Collectively, these extensions would improve the fidelity, generalizability, and
practical applicability of leakage-free bandgap-prediction models.

\section{Conclusion}

In this study, a leakage-free machine-learning framework was developed for predicting
bandgaps using a physically curated subset of the JARVIS-DFT dataset. After removing
leakage-prone descriptors and enforcing strict data-cleaning criteria, the models were
evaluated across three progressively enriched feature sets. Despite expanding the
descriptor space from fundamental physical features to engineered descriptors and
finally to Matminer-derived descriptors, including some engineered attributes, the
predictive accuracy of classical ML models remained essentially unchanged, saturating
around $R^2 \approx 0.88$--0.90. SHAP analysis revealed a consistent phase-independent
behavior: the dielectric tensor components dominated the predictions, whereas the
chemical and mechanical features acted as secondary contributors.

These findings show that for leakage-free bandgap regression on the JARVIS-DFT data,
adding more descriptors does not necessarily enhance model performance once the
leakage is controlled. Instead, a small set of physically meaningful features governs
the predictability, and classical models rapidly reach an accuracy ceiling. This ceiling should be interpreted as a diagnostic bound imposed by the available
leakage-free descriptors in the JARVIS-DFT dataset, rather than as a fundamental
upper limit on bandgap prediction accuracy. This
highlights the importance of careful data curation, leakage detection, and
interpretable machine learning in the field of materials informatics. This work is
intended as a controlled diagnostic study of leakage and feature saturation in
classical ML models on JARVIS-DFT bandgap data, rather than as a comprehensive
benchmark across architectures or datasets. Future work involving larger multi-source
datasets, deep-learning architectures, and transfer-learning strategies may help
surpass the saturation observed in this study and move closer to experimentally
aligned bandgap predictions.
\bibliographystyle{unsrt}
\bibliography{references}

\end{document}